\documentclass[letterpaper]{article} 
\usepackage{aaai23}  
\usepackage{times}  
\usepackage{helvet}  
\usepackage{courier}  
\usepackage[hyphens]{url}  
\usepackage{graphicx} 
\usepackage{natbib}  
\usepackage{caption} 
\usepackage{xcolor} 
\usepackage{algorithm}
\usepackage{algorithmic}
\usepackage{mmstyle}
\usepackage{multirow}
\usepackage{booktabs}
\usepackage{subfigure}
\usepackage{graphicx}
\usepackage{color}
\usepackage{xcolor}

\definecolor{mygreen}{RGB}{85,130,52}
\definecolor{myblue}{RGB}{48,118,183}

\usepackage{newfloat}
\usepackage{listings}
\DeclareCaptionStyle{ruled}{labelfont=normalfont,labelsep=colon,strut=off} 
\floatstyle{ruled}
\newfloat{listing}{tb}{lst}{}
\floatname{listing}{Listing}
\title{Self-supervised Action Representation Learning from \\
Partial Spatio-Temporal Skeleton Sequences}
\author{
    Yujie Zhou\textsuperscript{\rm 1, 4},
    Haodong Duan\textsuperscript{\rm 3},
    Anyi Rao\textsuperscript{\rm 3},
    Bing Su\textsuperscript{\rm 1, 2\thanks{Corresponding author}},
    Jiaqi Wang\textsuperscript{\rm 4}
    \\
}
\affiliations{
    \textsuperscript{\rm 1}Gaoling School of Artificial Intelligence, Renmin University of China, \\
    \textsuperscript{\rm 2}Beijing Key Laboratory of Big Data Management and Analysis Methods,\\
    \textsuperscript{\rm 3}The Chinese University of Hong Kong,
    \textsuperscript{\rm 4}Shanghai AI Laboratory\\
    yujiezhou@ruc.edu.cn, anyirao@link.cuhk.edu.hk, \{dhd.efz, subingats, wjqdev\}@gmail.com 
}

\usepackage{bibentry}

\begin{document}

\maketitle


\begin{abstract}

Self-supervised learning has demonstrated remarkable capability in representation learning for skeleton-based action recognition.
Existing methods mainly focus on applying global data augmentation to generate different views of the skeleton sequence for contrastive learning. 
However, due to the rich action clues in the skeleton sequences, existing methods may only take a global perspective to learn to discriminate different skeletons without thoroughly leveraging the local relationship between different skeleton joints and video frames, which is essential for real-world applications.
In this work, we propose a \textbf{P}artial \textbf{S}patio-\textbf{T}emporal \textbf{L}earning (\textbf{PSTL}) framework to exploit the local relationship from a partial skeleton sequences built by a unique spatio-temporal masking strategy. 
Specifically, we construct a negative-sample-free triplet steam structure that is composed of an anchor stream without any masking, a spatial masking stream with Central Spatial Masking (CSM), and a temporal masking stream with Motion Attention Temporal Masking (MATM). 
The feature cross-correlation matrix is measured between the anchor stream and the other two masking streams, respectively.
(1) Central Spatial Masking discards selected joints from the feature calculation process, where the joints with a higher degree of centrality have a higher possibility of being selected. 
(2) Motion Attention Temporal Masking leverages
the motion of action and remove frames that move faster with a higher possibility. 
Our method achieves state-of-the-art performance on NTURGB+D 60, 
NTURGB+D 120 and PKU-MMD under various downstream tasks.
Furthermore, to simulate the real-world scenarios, a practical evaluation is performed where some skeleton joints are lost in downstream tasks.
In contrast to previous methods that suffer from large performance drops, our PSTL can still achieve remarkable results under this challenging setting, validating the robustness of our method.
Our code is available at \url{https://github.com/YujieOuO/PSTL.git}.
\end{abstract}

\section{Introduction}
\label{sec:introduction}


In recent years, human action recognition has been extensively studied for various applications in the real world, 
such as security, human-robot interaction, and virtual reality.
With the development of sensor technology~\cite{zhang2012microsoft,shotton2011real} and human pose estimation methods~\cite{Cao_2017_CVPR,hua2022weakly}, 
skeleton data becomes easier to obtain.  
Its robustness against the variation of appearance and backgrounds 
has attracted many researchers to explore fully supervised methods 
for action recognition, 
such as applying CNN~\cite{ke2017a}, 
RNN~\cite{du2015hierarchical} or GCN~\cite{yan2018spatial,shi2019two} on skeleton data. 


However, fully supervised learning relies on abundant manually annotated labels, 
which is time-consuming and expensive. 
Therefore, self-supervised frameworks, \eg, MoCo~\cite{he2020momentum}, 
SimCLR~\cite{chen2020simple}, Barlow Twins~\cite{zbontar2021barlow}, 
and MAE~\cite{he2022masked} have received a lot of attention. 
For self-supervised learning on skeleton data, 
early methods also apply pretext tasks~\cite{lin2020ms2l,su2020predict,zheng2018unsupervised}.
While nowadays, most researchers focus on contrastive learning and apply MoCo structure 
to the skeleton squences~\cite{thoker2021skeleton,Li_2021_CVPR,rao2021augmented,guo2022contrastive}.

The key of current contrastive learning-based methods is finding effective global data augmentation and using them to create various views of the skeleton carefully. 
However, due to the rich action clues stored in the skeleton sequence, this paradigm can readily discriminate different views from a global perspective without thoroughly leveraging the local relationship between different skeleton joints and video frames. 
As a result, it limits the performance and robustness of the model in real-world scenarios. Moreover, a large batch size or a memory bank is necessary for contrastive learning, which is unsuitable for a small amount of skeleton data. 
To evaluate the robustness of the current methods in real-world scenarios (such as video surveillance), 
we formulate a new evaluating setting where only a partial skeleton is available for downstream tasks to simulate that the skeleton data captured by the camera are shaded. 
In this challenging setting, previous works suffer severe performance drops. 
We conjecture that if the local spatial and temporal relationship can be explored and captured, it will boost the representation ability and robustness of the model on downstream tasks.

Therefore, we propose a Partial Spatio-Temporal Learning (PSTL) framework to exploit the local relationship from a partial skeleton sequence built by a unique spatio-temporal masking strategy. 
Specifically, we construct a negative-sample-free triplet steam structure that is composed of an anchor stream, a spatial masking stream with Central Spatial Masking (CSM), and a temporal masking stream with Motion Attention Temporal Masking (MATM). 

In each stream, an ordinary augmentation is first applied to improve the diversity of the input samples. 
(1) In the spatial masking stream, Central Spatial Masking discards selected joints from the feature calculation process. Notably, it is not simply set the selected joint position to zeros, which is unreasonable for skeleton data. 
Considering that the topology of the human skeleton is a predefined graph, the joints with a higher degree of centrality get stronger connectivity. By filtering out such joints with higher probabilities, the encoder can focus more on these skeletons that are less explored. 
(2) In the temporal masking stream, considering the frames that change quickly always contain more semantic information about the actions, 
we propose a Motion Attention Temporal Masking strategy, where
frames moves quickly are more likely to be masked.
(3) In the anchor stream, the masking policy is not introduced to keep the original semantic information. A shared encoder extracts 
the triple stream features simultaneously.
Inspired by the loss of Barlow Twins~\cite{zbontar2021barlow},
we adopt the cross-correlation matrix to 
capture the correlation between the anchor stream and the other two masking streams, respectively.

As a negative-sample-free self-supervised method, 
PSTL encourages the empirical cross-correlation matrix 
between learned representations to be an identity matrix, 
and minimizes the redundancy within the 
representation vector to alleviate feature collapse.
In this way, the spatial and temporal relationship can be well captured 
and the joint robustness can be boosted for downstream tasks. 
Extensive results on three large benchmark
demonstrate the effectiveness of our proposed masking strategy. 
Our contributions:

\vspace{2pt}
\begin{itemize}

\item We propose Partial Spatio-Temporal skeleton representation Learning (PSTL), a negative-sample-free self-supervised framework for skeleton-based action representation.

\item We are the first to introduce the masking strategy 
to skeleton-based self-supervised learning and 
using Central Spatial Masking (CSM) strategy during pre-training to 
improve the joint robustness of learned representation for downstream tasks.

\item A unique Motion Attention Temporal Masking strategy (MATM) 
is proposed to use the motion of skeleton data to guild 
temporal masking modules to find and mask key frames of the sequences. 

\item Extensive experiment results on NTU-60, NTU120, and PKU-MMD datasets 
show the proposed PSTL achieves the-state-of-art performance under diverse downstream tasks,
such as linear evaluation, finetune evaluation, and semi-supervised evaluation protocol.

\end{itemize}

\section{Related Work}
\label{sec:related}

{\noindent\bf Skeleton-based action recognition.}
To tackle skeleton-based action recognition,
early methods mainly focus on hand-crafted features~\cite{vemulapalli2014human,xia2012view}.
With the development of the deep neural network, 
skeleton data is processed as pseudo-images with CNNs~\cite{ke2017a,kim2017interpretable}.
Others structure it as a long-term sequence
and feed it into RNNs~\cite{du2015hierarchical,zhang2017view}.
Later on, ST-GCN~\cite{yan2018spatial} structures the skeleton data as a predefined spatial graph and 
utilize GCN to integrate the skeleton joint information. 
After that, many GCN-based methods are proposed, 
which contain more data stream or add attention mechanisms~\cite{shi2019two,song2020stronger,chen2021multi}. 
In this paper, we also utilize the widely used ST-GCN as the backbone.

{\noindent\bf Self-supervised representation learning.}
Self-supervised learning aims to acquire a high-level representation from a large number of unlabeled samples. 
Early methods rely on pretext tasks to generate supervision, 
such as jigsaw puzzles~\cite{noroozi2016unsupervised,wei2019iterative}, 
colorizing~\cite{larsson2016learning, zhang2016colorful}, 
or rotation prediction~\cite{gidaris2018unsupervised}. %
Then, discriminative methods like MoCo~\cite{he2020momentum} and SimCLR~\cite{chen2020simple} achieve 
remarkable performance, 
which are comparable to supervised learning. 
By distinguishing the positive pairs from the negative ones, 
the backbone network can acquire the representation with rich discriminant information. 
But it requires a large batch size or memory bank to store the negative samples. 
To tackle this issue, negative-sample-free methods are introduced 
such as BYOL~\cite{grill2020bootstrap}, SimSiam~\cite{chen2021exploring}, and Barlow Twins~\cite{zbontar2021barlow}. 
They do not require explicit construction of negative sample pairs,
but most of them utilize an asymmetric network to avoid feature collapse.
Specially, 
Barlow twins avoids the reliance on the symmetry-breaking network 
but minimizes the redundancy within the representation vector to alleviate collapse. 
More recently, mask auto-encoder~\cite{he2022masked} is introduced to learn effective visual representations
from local complementary information.

{\noindent\bf Self-supervised skeleton-based action recognition.}
To apply self-supervised learning to skeleton data,
LongT GAN~\cite{zheng2018unsupervised} uses a recurrent encoder-decoder GAN to reconstruct the input sequence.
Based on this, Predict\&Cluster~\cite{su2020predict} proposes
a decoder to improve the representation ability of the encoder. 
MS2L~\cite{lin2020ms2l} proposes a multi-task self-supervised learning framework,
including motion prediction and jigsaw puzzle. 
AS-CAL~\cite{rao2021augmented} uses momentum LSTM to regularize the feature space together with several skeleton augmentation strategy.
ISC~\cite{thoker2021skeleton} proposes inter-skeleton contrastive method that combines sequence-based and graph-based representation.
CrossCLR~\cite{Li_2021_CVPR} introduces
SkeletonCLR that applies memory bank to store the negative samples and 
a cross-view knowledge mining strategy to capture more comprehensive representation. 
AimCLR~\cite{guo2022contrastive} aims to utilize plenty of augmentation to introduce movement patterns and force 
the encoder to learn more general representations. 
However, these contrastive-based methods heavily rely on strong data augmentation strategies, 
but the redundancy within the spatial joints and temporal frames have not been considered, 
which can improve the robustness of 3D visual representation and essential for downstream task.
Hence, we introduce PSTL and propose a spatial-temporal mask strategy to find a more general representation. 
\section{Method}
\label{sec:method}

\begin{figure*}[!t]
\begin{center}
\includegraphics[width=0.95\textwidth]{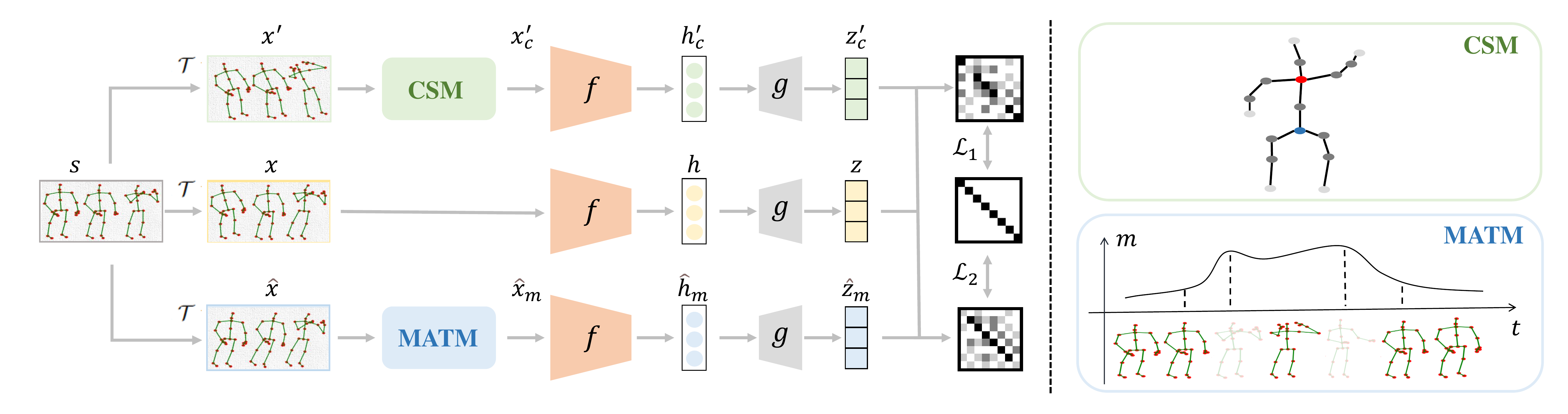} 
\vspace{-6pt}
\caption{The overall framework of Partial Spatio-Temporal skeleton representation Learning. 
In Central Spatial Masking \textbf{(CSM)},  the degree is red (4), blue (3), dark grey (2) and light grey (1).
In Motion Attention Temporal Mask \textbf{(MATM)}, 
$m$ denotes motion density while $t$ is the time code.
}
\label{fig:PSTL}
\end{center}
\vspace{-10pt}
\end{figure*}

As shown in Fig~\ref{fig:PSTL},
Partial Spatio-Temporal skeleton representation Learning (PSTL) 
takes a triple stream architecture, which is based on our Skeleton Barlow Twins (SkeletonBT),
and applies Central Spatial Masking (CSM) on its spatial masking stream and Motion Attention Temporal Masking (MATM) on its temporal masking stream together with an extra anchor stream to keep the original semantic information. 
Specifically,
1) \textbf{SkeletonBT} is a basic negative-sample-free double stream network
with ordinary augmentation for skeleton self-supervised learning. 
2) \textbf{CSM} first adjusts the computing mode of the GCN backbone
to ensure the information of masked joints out of the feature calculation process. 
Then, the joints with higher degree centrality are assigned higher probability to be masked. 
3) \textbf{MATM} calculates the corresponding motion value for each frame of the action, 
which serves as the attention weight to select the masked frames. 

\subsection{Skeleton Barlow Twins}
\label{subsec:Bt}

Inspired by the recent advance on negative-sample-free self-supervised methods~\cite{zbontar2021barlow} that avoid negative sample 
and do not need a large batch size/memory bank, 
we develop Skeleton Barlow Twins (SkeletonBT) to learn the skeleton-based action 
representation.
It is a symmetric network, 
where a shared encoder is used to generate features.

\begin{figure}[!t]
  \centering
  \includegraphics[width=0.95\linewidth]{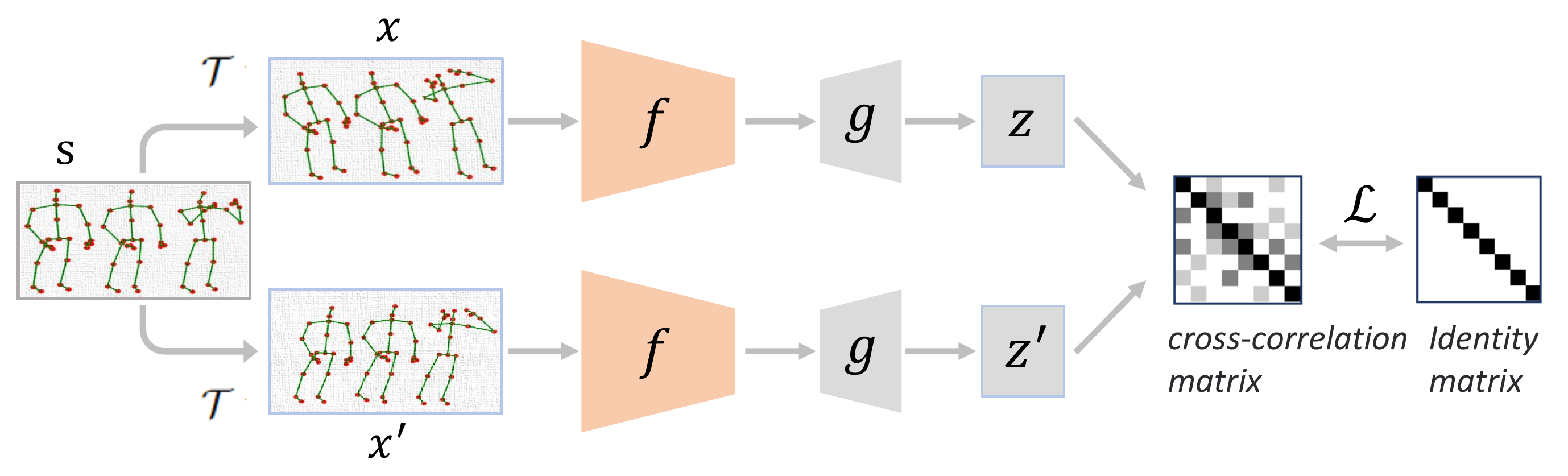}  
  \vspace{-5pt}
  \caption{The framework of Skeleton Barlow Twins.}
  \label{fig:SkeletonBT}
  \vspace{-5pt}
\end{figure}

Specifically, a 3D human skeleton sequence can be denoted as
$s \in \mathbb{R}^{C \times T \times V}$ ,
which has $T$ frames and $V$ joints. 
$C$ is the channel dimension representing the 3D position. 
As shown in Fig.~\ref{fig:SkeletonBT},
the input skeleton $s$ is first applied with an 
ordinary augmentation $\mathcal{T}$
to get diverse views $x$ and $x^{'}$. 
The encoder $f$ is used to extract features: 
$h = f(x;\theta)$ and $h^{'} = f(x^{'};\theta)$, 
where $h, h^{'} \in \mathbb{R}^{c_{h}}$. 
Then a projector $g$ maps each feature 
to the higher dimension space and acquire the embeddings: 
$z = g(h)$ and $z^{'} = g(h^{'})$, 
where $z, z^{'} \in \mathbb{R}^{c_{z}}$. 
At last, by encouraging the empirical cross-correlation matrix 
between embeddings of the distorted variations be an identity matrix,
the encoder can capture the relationship between two streams. 
The loss can be formulated as: 
\begin{equation}
\mathcal{L} = \sum_{i}\left(1-\mathcal{C}_{i i}\right)^{2}+\lambda \sum_{i} \sum_{j \neq i} \mathcal{C}_{i j}^{2}.
\label{eq:BT}
\end{equation}
Here $\mathcal{C}$ is the cross-correlation matrix 
computed between embeddings $z$ and $z^{'}$ 
along the batch dimension $b$: 

\begin{equation}
\label{eq:cc}
\mathcal{C}_{i j} = \frac{\sum_{b} z_{b, i} z_{b, j}^{'}}{\sqrt{\sum_{b}\left(z_{b, i}\right)^{2}} \sqrt{\sum_{b}\left(z_{b, j}^{'}\right)^{2}}},
\end{equation}
where $i,j$ represent the embedding dimension. 

Specifically, in Eq.~\eqref{eq:BT} the first term encourages the diagonal elements of $\mathcal{C}$ to 1,
which makes the embedding invariant to the variety augmentation applied. 
The second term equates the off-diagonal elements of $\mathcal{C}$ to 0, 
helping the different embedding components to be decoupled with each other, 
which can minimize the redundancy within the representation.
$\lambda$ serves as a trade-off parameter to balance two terms.

\subsection{Central Spatial Masking}
\label{subsec:CSM}
To boost the joint robustness of learned representation,
we propose a Central Spatial Masking (CSM) module that 
encourages partial skeleton data to generate similar features as the whole skeleton data. 
The relationship between masked and unmasked joints can be captured by the encoder, 
which is essential for the hard downstream task.

As discussed above, it is critical to construct partial skeleton data for the encoder. 
Early methods~\cite{rao2021augmented} try to directly set the values of masked joints to zero.
But it is unreasonable for skeleton data since
the joint semantic information is the 3D position that is changed. 
For CSM, we first adjust the computing mode of the backbone ST-GCN~\cite{yan2018spatial}.
Specifically, for a skeleton sequence $s$, the masked joints are abandoned and then
drop the corresponding rows and columns in the adjacency matrix, 
forcing masked joints out of the calculation process and the learned feature 
is only generated by the unmasked ones. 

Next, we take the advantage of the degree centrality 
of the human skeleton graph topology, 
and observe that the joints with more degrees can 
acquire richer neighborhood information. 
Thus, we assign higher probabilities to mask 
such connective joints which help the encoder capture the relationship between 
a wider range.
As shown in the green box of Fig~\ref{fig:PSTL}, the human skeleton joint is $V_i, i \in (1,2,...,n)$,
where $n$ is the total number of joints. 
Note that there are four types of joints.
The degree of light gray joints located at the margin of the graph is 1. 
The degree of dark gray joints is 2, which are more connective 
and their number is the majority in the total graph. 
The blue one and red one hold degree of 3 and 4, 
which are the center of the upper half and lower half of the body. 
Masking such center joints can help the encoder to
capture the relationship between a large range of joint information.
To this end, our CSM strategy first calculates the degree $d_i$ of each joint $V_i$, 
and sets their masked probability:
\begin{equation}
p_i = \frac{d_i}{\sum_{j=1}^{n} d_j}.
\end{equation}


\subsection{Motion Attention Temporal Masking}
\label{subsec:MATM}
Compared with the spatial joints, the temporal information is more redundant and
the semantic of action is more concentrated. 
Some key-frames play an pivotal role in representing the semantics of an action. 
To utilize this feature, 
our MATM aims to find such key-frames and mask them to 
generate a hard pair for anchor stream and temporal masking stream in our proposed PSTL. 
By encouraging the empirical cross-correlation matrix between embeddings of two stream
be an identity matrix, the encoder can capture the relationship between 
redundant frames and key-frames effectively. 

Specifically, the motion $m \in \mathbb{R}^{C \times T \times V}$ of the sequence $s$
is computed by the temporal displacement between frames:
$m_{:, t, :} = x_{:, t+1,:}-x_{:, t, :}$. 
Then we acquire the overall motion rate of a frame which serves as the attention weight:
\begin{equation}
a_t = \frac{m_t^2} {\sum_{i=1}^{T} {m_i^2}}.
\end{equation}
Then, the top-$K$ attention weights $a_{t_1},...,a_{t_K}$ are selected 
and the sequence $x_{t_1},...,x_{t_K}$ serves as the key-frames which
contain more semantic information about the actions. 
As shown in the blue box of Fig~\ref{fig:PSTL}, the bottom one is a skeleton sequence of \textit{put on glasses}, 
the top one shows the motion density. 
We can find that some frames in the middle of the sequence acquire higher motion values, 
which are the key-frames for this action. 
Thus, we mask these key-frames and encourage the feature from masked sequence close to 
the anchor feature which contains total semantic information. 

\subsection{Loss Function}
\label{subsec:PSTL}
Recall the pipeline of the proposed triple stream PSTL shown in Fig.~\ref{fig:PSTL}.
Given a skeleton sequence $s$, an ordinary augmentation is applied 
to obtain different views $x,x^{'},\hat{x}$. 
For sequence $x^{'}$, a Central Spatial Masking strategy is used to 
generate partial skeleton data $x_c^{'}$. 
Considering the degree centrality of the skeleton graph, 
the joints with more degrees are assigned higher masked probability in this module.   
For sequence $\hat{x}$, a Motion Attention Temporal Masking strategy is applied.
We first compute the motion value of each frame which represents the key 
fragment in a skeleton action. 
Then the top-$K$ frames are selected to be masked 
and to further improve diversity,
$K$ random frames are still masked from remain sequence to 
generate $\hat{x}_m$. 
The sequence $x$ in the anchor stream does not introduce any mask
strategy to keep the original semantic information. 

After that, the shared encoder $f$ is used to extract the features simultaneously.
$h = f(x;\theta), h_c^{'} = f(x^{'};\theta), \hat{h}_m = f(\hat{x}_m;\theta)$,
where $\theta$ is the parameters learned by the encoder
which is also used in downstream tasks. 
Then a projector $g$ is applied and the features are projected to
the embeddings $z, z_c^{'}, \hat{z}_m$ in a high dimension space. 
\ie $z = g(h), z_c^{'} = g(h_c^{'}), \hat{z}_m = g(\hat{h}_m)$. 
To capture the relationship between masked joints and unmasked ones, 
we compute the cross-correlation matrix $\mathcal{C^{'}}$ between embeddings $z$ and $z_c^{'}$. 
The calculation process refers to Eq.~\eqref{eq:cc}.
And then $\mathcal{L}_1$ is applied:

\begin{equation}
\label{eq:BT1}
\mathcal{L}_1 = \sum_{i}\left(1-\mathcal{C^{'}}_{i i}\right)^{2}+
\lambda \sum_{i} \sum_{j \neq i} \mathcal{C^{'}}_{i j}^{2},
\end{equation}
where the first term encourages the diagonal element of $\mathcal{C^{'}}$ to 1.
It forces the representation of partial data to be similar to the total one. 
The second term is designed to prompt the different embedding components to be decoupled with each other,
which helps to minimize the redundancy within the representation and avoid it being a constant. 

Similarly, the cross-correlation matrix $\mathcal{\widehat{C}}$ between $z$ and $\hat{z}_m$
is acquired for the $\mathcal{L}_2$:
\begin{equation}
\label{eq:BT2}
\mathcal{L}_2 = \sum_{i}\left(1-\mathcal{\widehat{C}}_{i i}\right)^{2}+
\lambda \sum_{i} \sum_{j \neq i} \mathcal{\widehat{C}}_{i j}^{2},
\end{equation}
which is utilized to capture the relationship between masked and unmasked frames.
Note that the trade-off parameter $\lambda$ is used to balance the dimension difference 
between first and second terms, so it keep same on Eq.~\eqref{eq:BT1} and Eq.~\eqref{eq:BT2}. 

The total loss $\mathcal{L}_p$ for our PSTL can be formulated as:

\begin{equation}
\mathcal{L}_p = \mathcal{L}_1 + \mathcal{L}_2.
\end{equation}

\section{Experiments}
\label{sec:experiment}

\subsection{Datasets}

\noindent\textbf{NTU-RGB+D 60.}
NTU-60 dataset~\cite{shahroudy2016ntu} is collected by Microsoft Kinect sensors. 
It contains $56,578$ skeleton sequences of $60$ action categories,
performed by $40$ volunteers.
The skeleton of each subject is represented by $25$ joints, 
while each video has no more than two subjects. 
There are two official dataset splits: 
1) Cross-Subject (xsub): half of the subjects belong to the training set, 
and the rest make up the testing sets;
2) Cross-View (xview): training and testing sets are captured by cameras with different views. 

\noindent\textbf{NTU-RGB+D 120.}
NTU-120 dataset~\cite{liu2019ntu} is the extended version of the NTU-60,
which contains $113,945$ skeleton sequences of $120$ action categories, 
performed by $106$ volunteers. 
There are two official dataset splits:
1) Cross-Subject (xsub): the training set is performed by $53$ subjects,
and testing data is performed by the rest volunteers;
2) Cross-Setup (xset): the training and testing sets are captured by cameras with different setups. 

\noindent\textbf{PKU-MMD.}
PKU-MMD dataset~\cite{liu2020benchmark} is captured via 
the Kinect v2 sensors from multiple viewpoints.
It has almost $20000$ action instances in $51$ categories collected by $66$ subjects. 
There are two parts of the dataset:
1) Part I contains $21539$ sequences;
2) Part II contains $6904$ sequences.
Part II is more challenging due to the large view variation with more skeleton noises.

\subsection{Implementation Details}

To perform an apple-to-apple comparison with other methods, 
we follow the same pre-processing methods of CrossCLR~\cite{Li_2021_CVPR} and AimCLR~\cite{guo2022contrastive}, 
which resize the skeleton sequences to 50 frames. 
Similarly, ST-GCN~\cite{yan2018spatial} with 16 hidden channels is used as the backbone.
For all experiments (both representation learning and downstream tasks), we adopt Adam optimizer and
the CosineAnnealing scheduler with 150 epochs. 
The mini batch size is 128. 

\noindent\textbf{Data Augmentation.}
Before applying the joint and temporal masking strategy proposed, 
some ordinary augmentations are used to generate different views
of the input skeleton sequences.
The ordinary augmentations include 3 spatial augmentations and 1 temporal augmentation. 

(1) \textit{Shear.}
It is a linear transformation for skeleton 3D coordinates.
A shear matrix is
$$
\mathbf{S}=\left[\begin{array}{ccc}
1 & s_{12} & s_{13} \\
s_{21} & 1 & s_{23} \\
s_{31} & s_{32} & 1
\end{array}\right]
,
$$ 
where $s_{12}, s_{13}, s_{21}, s_{23}, s_{31}, s_{32}$ are the shear factors and are
randomly sampled from $[-\beta \sim \beta]$. $ \beta $ is the amplitude to adjust 
the augmentation strength. 
Here we set $ \beta = 1$.  
By directly multiplying the shear matrix with the 3D coordinates, the human pose slants
in a random angle. 

(2) \textit{Crop.}
As a temporal augmentation, Crop aims to improve data diversity 
by first padding part of frames to the original sequence and then 
randomly cropping it to the original length. 
The padding length is $\gamma\, T$, $T$ is the original sequence length, and $\gamma$ is the padding ratio (we set $\gamma = 1 / 6$). 

(3) \textit{Rotate.}
This is an efficient spatial augmentation which 
helps the representation adapt to the variation of spatial perspective. 
Specifically, it first randomly selects an axis $M \in \{X, Y, Z\} $ as main axis and then a random angle in $[0 \sim \pi/6]$ is selected for it to rotate. 
Next, the rest two axis both randomly choose the angle in $[0 \sim \pi/180]$.

(4) \textit{Spatial Flip.}
It is another spatial augmentation,
which swaps the left and right sides of the skeleton data with the probability $p=0.5$. 

\subsection{Experiment Setting}

\noindent\textbf{Self-supervised Pre-training.}
We use ST-GCN~\cite{yan2018spatial} as the encoder,
which extracts $256$-dim features from skeleton sequences.
Then we project features to $6144$-dim embeddings. 
$\lambda$ in the loss of each stream is set to 2e-4. 
A 10-epoch warmup is used for stabilizing the training process. 
The weight decay is set to 1e-5.
Note that for all datasets and evaluation settings, 
the number of mask joints is 9 in CSM and $K=10$ in MATM.
To compare with the state-of-the-art, we use multiple skeleton modalities,
including joint (\textbf{J}), bone (\textbf{B}), and motion (\textbf{M}).
We evaluate the learned representations with various evaluation protocols.

\noindent\textbf{Linear Evaluation.}
We add a linear classifier on top of the frozen pre-trained encoder,
and then train the recognizer on the target skeleton action recognition dataset with initial learning rate $0.01$.

\noindent\textbf{Partial Body Evaluation.}
To encourage the learned representation to adapt to real-life scenarios 
with skeleton shaded or partial body information lost,
we introduce the Partial Body Evaluation protocol.
Basically, we follow the settings of Linear Evaluation. 
However, some joints or body parts may be masked to simulate the joint/part missing in extracted skeletons.
We consider two settings:
1) Joint shaded: some joints are masked randomly during evaluation;
2) Body part shaded: we divide skeletons into 5 major parts\footnote{left/right arms, left/right legs, torso} and randomly drop parts during evaluation.


\noindent\textbf{Finetuning.}
We append a linear classifier on top of the encoder and then finetune the entire network
on the target skeleton action recognition dataset with an initial learning rate $0.005$. 

\noindent\textbf{Semi-Supervised Learning.}
This protocol shares the same setting as ``Finetuning",
but only uses $1\%$ or $10\%$ training labeled data that is randomly selected.

\begin{table*}[!ht]
\begin{center}
\begin{tabular}{l|c|cc|cc|cc|cc|cc|cc}
\toprule
\multicolumn{1}{l|}{\multirow{3}{*}{Method}} &
  \multicolumn{1}{c|}{\multirow{3}{*}{Stream}} &
  \multicolumn{4}{c|}{NTU-60(\%)} &
  \multicolumn{4}{c|}{PKU(\%)} &
  \multicolumn{4}{c}{NTU-120(\%)} \\ \cline{3-14} 
\multicolumn{1}{c|}{} &
  \multicolumn{1}{c|}{} &
  \multicolumn{2}{c|}{xsub} &
  \multicolumn{2}{c|}{xview} &
  \multicolumn{2}{c|}{part I} &
  \multicolumn{2}{c|}{part II} &
  \multicolumn{2}{c|}{xsub} &
  \multicolumn{2}{c}{xset} \\
\multicolumn{1}{c|}{} &
  \multicolumn{1}{c|}{} &
  acc. &
  \multicolumn{1}{c|}{$\Delta$} &
  acc. &
  \multicolumn{1}{c|}{$\Delta$} &
  acc. &
  \multicolumn{1}{c|}{$\Delta$} &
  acc. &
  \multicolumn{1}{c|}{$\Delta$} &
  acc. &
  \multicolumn{1}{c|}{$\Delta$} &
  acc. &  $\Delta$ \\\midrule
SkeletonBT    & \textbf{J} & 68.5 &       & 71.6 &       & 84.7 &       & 38.2 &       & 53.9 &       & 52.7 &  \\
\textbf{PSTL}          & \textbf{J} & \textbf{77.3} & +8.8  & \textbf{81.8} & +10.2  & \textbf{88.4} & +3.9  &\textbf{ 49.3} & +11.1 & \textbf{66.2} & +12.3 & \textbf{67.7} & +15.0 \\\midrule
SkeletonBT    & \textbf{M} & 58.9 &       & 58.2 &       & 63.2 &       & 30.3 &       & 43.5 &       & 44.7 &   \\
\textbf{PSTL}          & \textbf{M} & \textbf{69.9} & +11.0 & \textbf{72.6} & +14.4 & \textbf{80.4} & +17.2 & \textbf{34.7} & +4.4  & \textbf{65.4} & +21.9 & \textbf{57.7} & +13.0 \\\midrule
SkeletonBT    & \textbf{B} & 62.4 &       & 64.7 &       & 79.1 &       & 29.8 &       & 52.4 &       & 53.6 &   \\
\textbf{PSTL }         & \textbf{B} & \textbf{75.4} & +13.0 & \textbf{79.2} & +14.5 & \textbf{84.8} & +5.7  & \textbf{41.3} & +11.5 & \textbf{58.7} & +6.3  & \textbf{66.6} & +13.0 \\\midrule
3s-SkeletonBT & \textbf{J+M+B} & 69.3 &      & 72.2 &       & 84.8 &      & 44.1 &      & 56.3 &       & 56.8     &   \\
\textbf{3s-PSTL}       & \textbf{J+M+B} & \textbf{79.1} & +9.8 & \textbf{82.6} & +10.4 & \textbf{89.2} & +4.4 & \textbf{52.3} & +8.2 & \textbf{69.2} & +12.9 & \textbf{70.3} & +13.5  \\\bottomrule
\end{tabular}
\vspace{-5pt}
\caption{Linear evaluation results compared with SkeletonBT on NTU-60, PKU-MMD, and NTU-120 datasets.
"$\Delta$" represents the gain compared to SkeletonBT using the same stream data. 
\textbf{J}, \textbf{M} and \textbf{B} indicate joint stream, motion stream, and bone stream.}
\label{tab:compBT}
\vspace{-8pt}
\end{center}
\end{table*}

\begin{table}[!t]
\begin{center}
\resizebox{0.9\linewidth}{!}{
\begin{tabular}{ccc|cc}
\toprule
\multirow{2}{*}{SkeletonBT} & \multirow{2}{*}{w/ CSM} & \multirow{2}{*}{w/ MATM} & \multicolumn{2}{l}{NTU-60(\%)} \\
  &   &   & xsub & xview \\ \midrule
\checkmark &            &            & 68.5 & 71.6  \\
\checkmark & \checkmark &            & 71.8 & 75.4  \\
\checkmark &            & \checkmark & 73.7 & 78.0  \\
\checkmark & \checkmark & \checkmark & \textbf{77.3} & \textbf{81.8}  \\ \bottomrule
\end{tabular}
}
\vspace{-5pt}
\caption{Ablation studies of PSTL on NTU-60 dataset.}
\label{tab:overall}
\vspace{-9pt}
\end{center}
\end{table}


\subsection{Ablation Studies of PSTL}
We ablate the proposed CSM and MATM strategies on our baseline SkeletonBT respectively to 
show their effectiveness. 
The linear evaluation results are shown in Tab.~\ref{tab:overall}, 
SkeletonBT achieves 68.5\% and 71.6\% accuracy on xsub and xview.
We find that the CSM module improves about 3.3\%, 3.8\% Top-1 accuracies
and the MATM module brings about 5.2\%, 6.4\% performance gain on
xsub and xview, respectively. 
With the help of both modules, 
our PSTL achieves 8.8\% and 9.4\% increments. 
These results show that our proposed CSM and MATM strategies can 
encourage the encoder to acquire stronger features
which are more suitable for downstream tasks.

\noindent\textbf{Comparisons with SkeletonBT.}
To further verify the effectiveness of our proposed method, 
we conduct experiments on three datasets to 
compare our PSTL with the baseline method SkeletonBT in detail. 
As shown in Tab.~\ref{tab:compBT}, PSTL performs much better than SkeletonBT 
on all three different streams. 
For the fusion results, 3s-PSTL substantially improves
SkeletonBT, especially for the NTU-120 datasets. 

\noindent\textbf{Comparisons with Random Masking Strategy.}
To demonstrate the effectiveness of our masking strategy, 
we compare our CSM and MATM modules with random spatial masking (RSM) strategy 
and random temporal masking (RTM) strategy, which selects the mask position randomly. 
The linear evaluation results are shown in Tab.~\ref{tab:random}, 
with the help of our triple stream network, 
the random masking can capture the relationship between 
joints and frames and get a good results on downstream tasks. 
And then, our CSM and MATM can further strengthen this capacity 
to get a better accuracy.

\begin{table}[!t]
\begin{center}
\resizebox{0.56\linewidth}{!}{
\begin{tabular}{l|cc}
\toprule
\multirow{2}{*}{Method} & \multicolumn{2}{c}{NTU-60(\%)} \\
                        & xsub           & xview          \\ \midrule
RSM + RTM                 & 75.6           & 79.7           \\
RSM + MATM                & 76.4           & 81.2               \\
CSM + RTM                 & 76.1           & 80.4                \\
CSM + MATM                & \textbf{77.3} & \textbf{81.8}          \\ \bottomrule
\end{tabular}
}
\vspace{-5pt}
\caption{Compared with random masking strategy on NTU-60 dataset. 
RSM means random spatial masking and RTM means random temporal masking.}
\label{tab:random}
\vspace{-15pt}
\end{center}
\end{table}

\subsection{Comparison with State-of-the-art}

\noindent\textbf{Partial Body Evaluation.}
To verify the \emph{joint robustness} of extracted features,
we conduct the partial body evaluation on NTU-60 xsub datasets.
The first level of the evaluation protocols is 
\textbf{joint shaded}, 
which applies random masks on joints to simulate the situation of 
joints shade in realistic scenarios.
Tab.~\ref{tab:jointshad} shows that, without CSM and MATM, the baseline  SkeletonBT suffers
serious performance loss. 
With the number of mask joints increasing to 10, 
the SOTA method AimCLR~\cite{guo2022contrastive} 
and its baseline methods SkeletonCLR~\cite{Li_2021_CVPR} also 
suffer about 20.0\% and 17.6\% performance decrease. 
Note that the feature extracted by our PSTL acquires strong joint robustness,
and only 3.5\% performance decrease under 10 masked joints. 

The second level evaluation protocols \textbf{body part shaded}, as shown in Tab.~\ref{tab:partshad}, 
is much harder since we mask the total parts of the body.
But our PSTL still achieves better results than other methods.
To sum up, more skeleton sequence information is lost, 
and the advantage of our method becomes more obvious. 
And the results tell that with the help of
local relationship between joints and frames, our CSM and MATM strategy 
can improve the robustness of learned features, 
which is essential to downstream tasks.

\begin{table}[!t]
\begin{center}
		\resizebox{\linewidth}{!}{
\begin{tabular}{l|cccccc}
\toprule
\# of masked joints      & 0 & 2 & 4 & 6 & 8 & 10 \\ \midrule
SkeletonCLR  & 68.3 & 65.6  & 61.9  & 58.6  & 54.5  & 50.7   \\
SkeletonBT     & 68.5 & 65.0  & 61.7  & 58.6  & 55.0  & 50.6   \\
AimCLR       & 74.3 & 70.4  & 67.2  & 63.0  & 58.5  & 54.3  \\
\textbf{PSTL(ours)}  & \textbf{77.3} & \textbf{76.9}  & \textbf{76.2} & \textbf{75.7}  & \textbf{74.6}  & \textbf{73.8}  \\ \bottomrule
\end{tabular}
}
\vspace{-5pt}
\caption{Linear evaluation results of \textbf{joints shaded} on NTU-60 xsub dataset.}
\label{tab:jointshad}
\vspace{-5pt}
\end{center}
\end{table}

\begin{table}[!t]
\begin{center}
\resizebox{0.95\linewidth}{!}{
\begin{tabular}{l|ccccc}
\hline
\# of masked body parts       & 0 & 1 & 2 & 3 & 4 \\ \hline
SkeletonCLR  & 68.3 & 63.6 & 57.7  & 47.5  & 33.5 \\
SkeletonBT   & 68.5 & 64.0 & 58.5 & 50.2 & 38.4  \\
AimCLR       & 74.3 & 69.3 & 63.9 & 53.7 & 37.9 \\
\textbf{PSTL (ours)}  & \textbf{77.3} & \textbf{73.6} & \textbf{68.3} & \textbf{60.0} & \textbf{47.3} \\ \hline
\end{tabular}
}
\vspace{-5pt}
\caption{Linear evaluation results of \textbf{body part shaded} on NTU-60 xsub dataset.}
\label{tab:partshad}
\vspace{-15pt}
\end{center}
\end{table}

\noindent\textbf{Linear Evaluation.}
As shown in Tab.~\ref{tab:ntu60&120lp}, for a single stream, 
PSTL achieves considerable gains on all other methods on NTU60 and NTU120.
Compared with the current SOTA method AimCLR, 
our PSTL improves 
3.0\% and 2.1\% on Xsub and Xview respectively. 
The results indicate that the local relationship between joints and frames can 
help extract suitable features for downstream tasks. 
For the performance of the 3-streams, PSTL still 
achieves comparable results to current methods. 
Our method also achieves the best results on NTU-120 datasets. 
For single stream, PSTL outperforms the current SOTA method AimCLR 
2.8\% and 4.3\% on Xsub and Xset respectively. 
And the fusion can also achieve 1.0\% and 1.5\% increment.

\begin{table}[!t]
\begin{center}
\resizebox{0.95\linewidth}{!}{
\begin{tabular}{l|cc|cc}
\toprule
\multirow{2}{*}{Method}          & \multicolumn{2}{c|}{NTU-60(\%)} & \multicolumn{2}{c}{NTU-120(\%)} \\
                                 & xsub          & xview          & xsub           & xset           \\\midrule
\textit{Single-stream:}         &               &               &                &                \\
LongT GAN (AAAI 18)              & 39.1          & 48.1         &  -             &  -              \\
MS2L (ACM MM 20)                 & 52.6         & -             &  -             &  -              \\
AS-CAL (Inf Sci 21)             &  58.5         & 64.8          &  48.6          & 49.2           \\
P\&C (CVPR 20)                   & 50.7         & 76.3          &  42.7          & 41.7             \\
SeBiReNet (ECCV 20)              &  -            & 79.7         &   -            & -               \\
AimCLR (AAAI 22)                 & 74.3         & 79.7          &  63.4          & 63.4           \\
\textbf{PSTL (ours)}            & \textbf{77.3} & \textbf{81.8} & \textbf{66.2}& \textbf{67.7} \\\midrule
\textit{Three-stream:}          &               &               &                &                \\
3s-SkeletonBT (ours)            &   69.3        & 72.2          &  56.3          & 56.8          \\
3s-SkeletonCLR (CVPR 21)        &    75.0        & 79.8          &  60.7          &  62.6             \\
3s-Colorization (ICCV 21)       &   75.2        & 83.1          &  -             &   -             \\
3s-CrosSCLR (CVPR 21)           &  77.8        & 83.4          &  67.9          &  67.1              \\
3s-AimCLR (AAAI 22)             &   78.9        & 83.8          &  68.2          &  68.8               \\
\textbf{3s-PSTL (ours) }        &   \textbf{79.1}        &  \textbf{83.8}        & \textbf{69.2}         &  \textbf{70.3} \\\bottomrule
\end{tabular}
}
\vspace{-5pt}
\caption{Linear evaluation results on NTU-60 and NTU-120 datasets.}
\label{tab:ntu60&120lp}
\vspace{-10pt}
\end{center}
\end{table}

Noting that the PKU-MMD dataset has about half the data size of NTU-60,
Tab.~\ref{tab:pkulp} shows that the advantage of our method is more obvious in a small dataset. 
It is worth mentioning that our baseline method SkeletonBT can achieve better 
results than AimCLR on the single stream, 
which indicates that our negative-sample-free methods without a memory bank
is more efficient for small datasets.
Similarly, our PSTL achieves remarkable results on PKU-MMD datasets.
For single stream, PSTL outperforms AimCLR 5.0\% and 12.5\% on Part I and Part II.
For 3 stream fusion, our 3s-PSTL outperforms 3s-AimCLR 1.4\% and 13.8\% on both parts.

\begin{table}[!t]
\begin{center}
\resizebox{0.74\linewidth}{!}{
\begin{tabular}{l|c|c}
\toprule
Method                & Part I & Part II \\\midrule

\textit{Single-stream:}         &        &         \\
LongT GAN (AAAI 18)         &  67.7      & 26.0        \\
MS2L (ACM MM 20)             & 64.9      & 27.6        \\
SkeletonCLR (CVPR 21)       &  80.9      & 35.2        \\
ISC (ACM MM 20)             &  80.9     &  36.0        \\
AimCLR (AAAI 22)                & 83.4  & 36.8          \\
SkeletonBT (ours)               & 84.7  & 38.2         \\
\textbf{PSTL(ours) }            & \textbf{88.4}  & \textbf{49.3}    \\\midrule
\textit{Three-stream:}          &        &         \\
3s-CrosSCLR (CVPR 21)           & 84.9   & 21.2    \\
3s-AimCLR (AAAI 22)             & 87.8   & 38.5   \\
3s-SkeletonBT (ours)            & 84.8  & 44.1    \\
\textbf{3s-PSTL (ours)}         & \textbf{89.2}   & \textbf{52.3}   \\\bottomrule
\end{tabular}
}
\vspace{-5pt}
\caption{Linear evaluation results on PKU-MMD dataset.}
\label{tab:pkulp}
\vspace{-15pt}
\end{center}
\end{table}

\noindent\textbf{Finetune Evaluation.}
We compare PSTL with other self-supervised methods and
the supervised method ST-GCN using the same structure and parameters as ours. 
As shown in Tab.~\ref{tab:ntu60&120ft}, 
for the single steam, 
all methods use bone steam since 
the results are better than the joint stream. 
Our PSTL achieves better performance than others on 
NTU-60 and NTU-120 datasets.
Notably, only single steam PSTL outperforms supervised 3s-ST-GCN on NTU-120 datasets,
which indicates the effectiveness of our method. 
Similarly, 3s-PSTL outperforms 3s-CrossCLR and 3s-AimCLR 
for the fusion results on both datasets.  

\begin{table}[!t]
\begin{center}
\resizebox{0.95\linewidth}{!}{
\begin{tabular}{l|cc|cc}
\toprule
\multirow{2}{*}{Method}          & \multicolumn{2}{c|}{NTU-60(\%)} & \multicolumn{2}{c}{NTU-120(\%)} \\
                                 & xsub          & xview          & xsub           & xset           \\\midrule
\textit{Single-stream:}         &               &                &                &                \\
SkeletonCLR (CVPR 21)           &  82.2         & 88.9          & 73.6              & 75.3            \\
AimCLR (AAAI 22)                 & 83.0         & 89.2         &  77.2           & 76.0               \\
\textbf{PSTL (ours)}            &  \textbf{84.5}& \textbf{92.0 }& \textbf{78.6}   &  \textbf{78.9}     \\\midrule
\textit{Three-stream:}          &               &                &                &                \\
3s-ST-GCN (AAAI 18) ${}^{\S}$   &   85.2        & 91.4          &  77.2         & 77.1              \\
3s-CrosSCLR (CVPR 21)            &  86.2        & 92.5          & 80.5          & 80.4               \\
3s-AimCLR (AAAI 22)              &  86.9        & 92.8          & 80.1          & 80.9               \\
\textbf{3s-PSTL (ours) }         &  \textbf{87.1} &\textbf{93.9} & \textbf{81.3} & \textbf{82.6}   \\\bottomrule
\end{tabular}
}
\vspace{-5pt}
\caption{Finetune evaluation results on NTU-60 and NTU-120 datasets.
"§" means the model is trained
with a fully-supervised manner.}
\label{tab:ntu60&120ft}
\vspace{-5pt}
\end{center}
\end{table}

\noindent\textbf{Semi-Supervised Evaluation.}
To show the superiority of our approach,
we test all methods under the semi-supervised setting, where only 1\% or 10\% labeled data can be used such that models are more likely to overfit.
As shown in Tab.~\ref{tab:semi}, our PSTL far exceeds the current SOTA method AimCLR on PKU-MMD datasets.
And for fusion results, 3s-PSTL outperforms 3s-AimCLR 8.6\% on Part II. 
It demonstrates that such contrastive methods which need a large memory bank 
are subjected to serious performance loss with insufficient samples. 
But our negative-sample-free PSTL can well alleviate this problem and 
achieve better results.

\begin{table}[!t]
\begin{center}
\resizebox{0.75\linewidth}{!}{
\begin{tabular}{l|c|c}
\toprule
Method                & Part I & Part II \\\midrule
\textit{1\% labeled data:}     &        &         \\
LongT GAN (AAAI 18)   & 35.8   & 12.4     \\
MS2L (ACM MM 20)      & 36.4  & 13.0        \\
ISC (ACM MM 21)       & 37.7  & -        \\
3s-CrosSCLR (CVPR 21) & 49.7  & 10.2        \\
3s-AimCLR (AAAI 22)   & 57.5  & 15.1        \\
\textbf{3s-PSTL(ours)} & \textbf{62.5}  & \textbf{16.9} \\\midrule
\textit{10\% labeled data:}    &        &         \\
LongT GAN (AAAI 18)   &  69.5  & 25.7       \\
MS2L (ACM MM 20)      & 70.3  & 26.1        \\
ISC (ACM MM 21)       & 72.1  & -        \\
3s-CrosSCLR (CVPR 21) & 82.9  & 28.6    \\
3s-AimCLR (AAAI 22)   & 86.1  & 33.4        \\
\textbf{3s-PSTL (ours)} & \textbf{86.9} & \textbf{42.0}      \\\bottomrule
\end{tabular}
}
\vspace{-5pt}
\caption{Semi-supervised evaluation results on PKU-MMD dataset.}
\label{tab:semi}
\vspace{-7pt}
\end{center}
\end{table}
\section{Conclusion}
\label{sec:conclusion}

In this paper, we introduce a triple stream network named
Partial Spatio-Temporal skeleton sequences Learning framework
for self-supervised skeleton-based action recognition. 
Firstly, all streams apply an ordinary augmentation for producing a variety of skeleton views.
Secondly, one stream introduces a Central Spatial Masking strategy to force the selected 
joints out of the calculation of GCN backbone. 
The third  stream utilizes a Motion Attention Temporal Masking module to find 
the keyframes and mask them. 
At last, the feature cross-correlation matrix
is measured to acquire the correlation among each stream. 
Experiments show that such a masking strategy can boost the robustness of 
representations of skeleton joints, 
which is essential to downstream tasks. 
our PSTL achieves remarkable performance 
in three large skeleton datasets under various evaluation settings.
\section{Acknowledgements}
\label{sec:acknowledgements}
This work was supported by Shanghai AI Laboratory, 
the National Natural Science Foundation of China No. 61976206 and No. 61832017, 
Beijing Outstanding Young Scientist Program NO. BJJWZYJH012019100020098, 
Beijing Academy of Artificial Intelligence (BAAI), 
the Fundamental Research Funds for the Central Universities, 
the Research Funds of Renmin University of China 21XNLG05, 
and Public Computing Cloud, Renmin University of China.

\bibliography{aaai23}

\section{Appendix}

The supplementary materials include
some additional ablation studies on the different 
number of masked joints in CSM module and different $K$ in MATM module. 

{\noindent\bf Effects of the number of masked joints in CSM.}
Tab.~\ref{tab:jointnum} illustrates the effects of the number of masked joints on our PSTL.
On NTU-60 xsub datasets, PSTL can acquire a good 
performance (77.3\%) on a single stream with 9 masked joints,
so we choose to report these results in our paper.
Note that, the same number of masked joints may not suitable for all
the three streams, and when the number is 11, 
PSTL can acquire a better fusion result (79.6\%).

{\noindent\bf Effects of top-$K$ in MATM.}
Tab.~\ref{tab:topk} shows the effects of different 
choice of $K$ on NTU-60 xsub datasets. 
The number of masked joints is fixed to 9.
When the top 10 keyframes are masked, 
PSTL can achieve the best result (77.3\%). 

{\noindent\bf Effects of the dimension of the embeddings.}
In our proposed PSTL, the structure of the projector $g$ 
is a 3-layer MLP.
The first two layers are followed by a
batch normalization layer and ReLU nonlinearity. 
The output of the encoder is called the 'representations' and the output of the projector is the 'embeddings'.

To figure out the effects of the embedding dimension, 
we conduct an ablation study with different dimensions on NTU-60 dataset. 
As shown in Tab.~\ref{tab:dim}, 
6144-dim can achieve a better performance than others 
on both xsub and xview splits.

\begin{table}[!t]
\begin{center}
\begin{tabular}{l|cccc}
\toprule
\# of masked joints & 5 & 7 & 9 & 11 \\ \midrule
J     & 76.7  & 76.0  & \textbf{77.3}  & 75.4   \\
M     & 74.8  & 75.1  & \textbf{75.4}  & 75.2   \\
B     & 69.1  & 69.6  & 69.9  & \textbf{70.5}   \\
J+M+B & 79.4  & 78.9  & 79.1  & \textbf{79.6}   \\ \bottomrule
\end{tabular}
\vspace{-4pt}
\caption{Ablation studies of the number of masked joints in CSM on NTU-60 xsub dataset.}
\label{tab:jointnum}
\end{center}
\vspace{-4pt}
\end{table}

\begin{table}[!t]
\begin{center}
\begin{tabular}{l|cccc}
\toprule
$K$   & 6 & 8 & 10 & 12 \\ \midrule
J     & 75.3  & 76.8  & \textbf{77.3}  & 76.1   \\
M     & 73.9  & 74.5  & \textbf{75.4}  & 74.9   \\
B     & 68.5  & 68.8  & 69.9  & \textbf{70.0}   \\
J+M+B & 77.7  & 78.8  & \textbf{79.1}  & 78.9   \\ \bottomrule
\end{tabular}
\vspace{-4pt}
\caption{Ablation studies of top-$K$ in MATM on NTU-60 xsub dataset.}
\label{tab:topk}
\end{center}
\vspace{-4pt}
\end{table}

\begin{table}[!t]
\begin{center}
\begin{tabular}{l|cc}
\toprule
\multirow{2}{*}{dimension} & \multicolumn{2}{l}{NTU-60(\%)} \\
                           & xsub  & xview          \\ \midrule
4096                       &    76.7 &  81.3        \\
6144                       &    \textbf{77.3} &  \textbf{81.8}         \\
8196                       &    75.8 &  81.3        \\ \bottomrule
\end{tabular}
\vspace{-4pt}
\caption{Ablation studies of the embedding dimension on NTU-60 dataset.}
\label{tab:dim}
\end{center}
\vspace{-4pt}
\end{table}

\end{document}